\documentclass[lettersize,journal]{IEEEtran}
\usepackage{amsmath,amsfonts}
\usepackage{array}
\usepackage{textcomp}
\usepackage{url}
\usepackage{graphicx}
\usepackage{cite}
\usepackage{mathrsfs}
\usepackage{ulem}
\usepackage{url}
\usepackage{makecell}
\usepackage{booktabs} 
\usepackage{pifont} 
\usepackage[table]{xcolor}
\usepackage[switch]{lineno}
\usepackage[utf8]{inputenc}
\usepackage[T1]{fontenc}
\usepackage[colorlinks=true, urlcolor=blue, linkcolor=red]{hyperref}
\newcolumntype{L}[1]{>{\raggedright\arraybackslash}p{#1}}
\newcolumntype{C}[1]{>{\centering\arraybackslash}p{#1}}
\newcolumntype{M}[1]{>{\centering\arraybackslash}m{#1}}
\newcolumntype{R}[1]{>{\raggedleft\arraybackslash}p{#1}}
\usepackage{xcolor}
\usepackage{scalerel}
\usepackage{tikz}
\usetikzlibrary{svg.path}

\definecolor{orcidlogocol}{HTML}{A6CE39}
\tikzset{
  orcidlogo/.pic={
    \fill[orcidlogocol] svg{M256,128c0,70.7-57.3,128-128,128C57.3,256,0,198.7,0,128C0,57.3,57.3,0,128,0C198.7,0,256,57.3,256,128z};
    \fill[white] svg{M86.3,186.2H70.9V79.1h15.4v48.4V186.2z}
                 svg{M108.9,79.1h41.6c39.6,0,57,28.3,57,53.6c0,27.5-21.5,53.6-56.8,53.6h-41.8V79.1z M124.3,172.4h24.5c34.9,0,42.9-26.5,42.9-39.7c0-21.5-13.7-39.7-43.7-39.7h-23.7V172.4z}
                 svg{M88.7,56.8c0,5.5-4.5,10.1-10.1,10.1c-5.6,0-10.1-4.6-10.1-10.1c0-5.6,4.5-10.1,10.1-10.1C84.2,46.7,88.7,51.3,88.7,56.8z};
  }
}

\newcommand\orcidicon[1]{\href{https://orcid.org/#1}{\mbox{\scalerel*{
\begin{tikzpicture}[yscale=-1,transform shape]
\pic{orcidlogo};
\end{tikzpicture}
}{|}}}}

\usepackage{hyperref} 

\definecolor{indigo}{RGB}{50,80,200}

\usepackage{multirow}
\begin{document}
\title{RoadFormer+: Delivering RGB-X Scene Parsing through Scale-Aware Information Decoupling and Advanced Heterogeneous Feature Fusion}

\author{Jianxin Huang$^{\orcidicon{0009-0004-6558-9589}}$,~\IEEEmembership{Student Member,~IEEE}, Jiahang Li$^{\orcidicon{0009-0005-8379-249X}}$,~\IEEEmembership{Graduate Student Member,~IEEE}, Ning Jia, \\Yuxiang Sun$^{\orcidicon{0000-0002-7704-0559}}$, Chengju Liu$^{\orcidicon{0000-0001-7543-0855}}$, Qijun Chen$^{\orcidicon{0000-0001-5644-1188}}$,~\IEEEmembership{Senior Member,~IEEE}, and Rui Fan$^{\orcidicon{0000-0003-2593-6596}}$,~\IEEEmembership{Senior Member,~IEEE}

\thanks{$^{}$
        {\tt\small }}
}

\markboth{}{}
\maketitle
\begin{abstract}Task-specific data-fusion networks have marked considerable achievements in urban scene parsing. Among these networks, our recently proposed RoadFormer successfully extracts heterogeneous features from RGB images and surface normal maps and fuses these features through attention mechanisms, demonstrating compelling efficacy in RGB-Normal road scene parsing. However, its performance significantly deteriorates when handling other types/sources of data or performing more universal, all-category scene parsing tasks. To overcome these limitations, this study introduces RoadFormer+, an efficient, robust, and adaptable model capable of effectively fusing RGB-X data, where ``X'', represents additional types/modalities of data such as depth, thermal, surface normal, and polarization. Specifically, we propose a novel hybrid feature decoupling encoder to extract heterogeneous features and decouple them into global and local components. These decoupled features are then fused through a dual-branch multi-scale heterogeneous feature fusion block, which employs parallel Transformer attentions and convolutional neural network modules to merge multi-scale features across different scales and receptive fields. The fused features are subsequently fed into a decoder to generate the final semantic predictions. Notably, our proposed RoadFormer+ ranks first on the KITTI Road benchmark and achieves state-of-the-art performance in mean intersection over union on the Cityscapes, MFNet, FMB, and ZJU datasets. Moreover, it reduces the number of learnable parameters by 65\% compared to RoadFormer. Our source code will be publicly available at \url{mias.group/RoadFormerPlus}.
\end{abstract}

\begin{IEEEkeywords}
urban scene parsing, heterogeneous features, Transformer, convolutional neural network.
\end{IEEEkeywords}

\section{INTRODUCTION}
\subsection{Background}
\IEEEPARstart{S}{CENE} parsing, also known as semantic segmentation, is crucial for the safety of autonomous driving \cite{li2023roadformer,fan2022mlfnet}. With the widespread adoption of deep learning techniques, convolutional neural networks (CNNs) and Transformers have demonstrated significant performance improvements over traditional geometry-based models in various image segmentation tasks \cite{lu2019monocular,fan2019road,zhou2022mtanet,min2022orfd,fan2019pothole}. However, single-modal networks that rely solely on RGB images show limitations in handling challenging conditions such as poor illumination and adverse weather \cite{hazirbas2017fusenet,huang2022discriminative}. To tackle these problems, subsequent research has explored the integration of useful information provided by additional data modalities. Depth or surface normal information has been utilized to identify spatially continuous regions \cite{zhou2022canet,fan2020sneroadseg}, while thermal images have been employed to enhance object recognition robustness under poor lighting conditions \cite{ha2017mfnet}. Furthermore, polarization information has been used to improve segmentation performance for transparent and highly reflective objects \cite{xiang2021polarization}. Our recently proposed RoadFormer \cite{li2023roadformer} effectively extracts heterogeneous features from RGB images and surface normal information and fuses these features for robust urban scene parsing, demonstrating notable efficacy in freespace and road defect detection. However, RoadFormer still has several limitations, especially when handling other types/sources of data. Moreover, the large quantity of parameters leads to considerable hardware resource consumption, thus limiting its deployment on terminal devices.

\subsection{Existing Challenges and Motivation}
Most existing data-fusion networks use symmetric duplex encoders to extract heterogeneous features from multiple data sources and fuse them to provide a more comprehensive understanding of the environment \cite{fan2020sneroadseg, zhou2022mtanet, wu2024s, feng2024segmentation}. However, while prior arts \cite{fan2020sneroadseg, li2023roadformer, zhang2023cmx} have been developed to capture more discriminative features using these weight-separating duplex encoders, directly fusing these features may produce ambiguous features, thus negatively impairing the performance of scene parsing \cite{zhao2023cddfuse}. Additionally, the symmetric models with extensive parameters require more hardware resources for training, particularly when compared to networks that rely solely on RGB images \cite{lv2024context}. Therefore, exploring an efficient and effective heterogeneous feature encoding strategy remains an under-explored research area that deserves more attention.

In addition to the heterogeneous feature extraction strategy, the performance of a data-fusion network also depends on the manner in which these features are fused. To address this issue, recent works \cite{li2023roadformer,zhang2023cmx,li2023ccafusion,lv2024context} employ learnable feature fusion approaches, which significantly outperform traditional, non-discriminative fusion methods that rely solely on element-wise concatenation or summation \cite{fan2020sneroadseg, hazirbas2017fusenet}. For example, RoadFormer \cite{li2023roadformer} adopts a Transformer-based approach to effectively capture long-range dependencies within heterogeneous features. On the other hand, RDFNet \cite{park2017rdfnet} employs CNN-based modules to process multi-scale features, effectively extracting local spatial cues, such as edges and textures, within a relatively small receptive field. However, these methods typically employ single-branch feature fusion blocks, where features extracted from RGB images and additional data types (referred to as ``X'' data) are fused using convolutional layers or attention mechanisms. Such single-branch feature fusion strategies may not always effectively encode both local and global contexts simultaneously, limiting their capacity to fully exploit the advantages of multi-modal/source data fusion. Considering Transformers' remarkable capability in modeling long-range dependencies and CNNs' robustness in local feature extraction \cite{zhao2023cddfuse}, further research into combining CNNs' local features and Transformer's global representations through a dual-branch fusion design to enhance scene parsing is highly warranted.

Moreover, while task-specific networks such as RoadFormer demonstrate impressive performance in RGB-Normal road scene parsing, their applicability to more universal urban scene parsing tasks and their effectiveness in handling diverse data types remain limited. For instance, RoadFormer exhibits a significant performance drop on comprehensive scene parsing datasets, such as the KITTI Semantics \cite{abu2018augmented} and Cityscapes \cite{cordts2016cityscapes}, compared to existing state-of-the-art (SoTA) RGB-D/Normal methods. Additionally, it performs suboptimally when processing RGB-Thermal/Polarization data \cite{ha2017mfnet,xiang2021polarization}. It is urged to design a universal RGB-X data-fusion network that performs robustly across multiple data sources for urban scene parsing. 

\subsection{Novel Contributions}

To address the aforementioned limitations, we first design a more efficient hybrid feature decoupling encoder (HFDE). Given the correlation between RGB images and their corresponding X data, we first replace the duplex encoder with a weight-sharing backbone to reduce the number of learnable parameters. We then employ an asymmetric architecture that independently utilizes two global feature enhancers (GFEs) and two local feature extractors (LFEs) to decouple heterogeneous features, effectively modeling their inherent differences at various scales.
Subsequently, we introduce a robust dual-branch multi-scale heterogeneous feature fusion (MHFF) block to fuse heterogeneous features in parallel, ensuring a comprehensive integration of global and local features. The MHFF block utilizes Transformer-based and CNN-based modules to parallelly fuse and calibrate multi-scale features. Our proposed RoadFormer+, an upgraded version of RoadFormer, with all these innovative components incorporated, demonstrates superior performance over RoadFormer across four RGB-Normal scene parsing datasets, while reducing the learnable parameters by around 65\%. Furthermore, RoadFormer+ achieves SoTA performance in RGB-Normal, RGB-Thermal, and RGB-Polarization scene parsing, demonstrating its exceptional applicability across a broad range of RGB-X data-fusion scenarios.

Our contributions can be summarized as follows:

\begin{itemize}

\item We introduce HFDE, which consists of a weight-sharing backbone and two pairs of independent GFEs and LFEs, to extract heterogeneous features and effectively capture both the correlation and inherent differences between RGB images and X data.

\item We design a dual-branch MHFF block to capture both global and local features simultaneously. It seamlessly integrates Transformer-based and CNN-based modules, so as to utilize different receptive fields to achieve advanced heterogeneous feature fusion.

\item We propose RoadFormer+, a novel urban scene parsing approach with fewer parameters compared with RoadFormer, which achieves SoTA performance across multiple RGB-X scene parsing datasets.

\end{itemize}

\subsection{Article Structure}
The remainder of this article is organized as follows: In Sect. \ref{Sect.relate}, we review related works on urban scene parsing. In Sect. \ref{Sect.methodology}, we introduce our proposed RoadFormer+. In Sect. \ref{Sect.experments}, we present quantitative and qualitative experimental results and their corresponding analyses. Finally, in Sect. \ref{Sect.conclusion}, we conclude this work and discuss potential future directions.

\section{Related Work}
\label{Sect.relate}

\subsection{Single-Modal Scene Parsing}

Since the introduction of fully convolutional network (FCN) \cite{long2015fully}, various CNN-based scene parsing networks have been developed. For instance, pyramid scene parsing network (PSPNet) \cite{zhao2017pyramid} uses pyramid pooling to capture semantic information at multiple scales. DeepLabV3+ \cite{chen2018encoder} employs atrous convolutions with different dilation rates to enrich the contextual feature encoding across scales. Additionally, MobileNetV2 \cite{sandler2018mobilenetv2} adopts lighter architectures based on depth-wise separable convolutions to reduce model parameters to computational demands. In these CNN-based networks, each convolutional kernel processes only a local region of the image at a time. This local receptive field design enables CNNs to excel at extracting local features, such as edges and textures \cite{chen2017deeplab}.

Transformers have gained prominence in scene parsing tasks due to their exceptional global aggregation capabilities compared to CNNs \cite{strudel2021segmenter}. The attention mechanisms within Transformers allow each token to interact with all others simultaneously \cite{vaswani2017attn}. These interactions help achieve a comprehensive understanding of the correlation between each token and the global context, thereby better extracting global features. Segmentation Transformer (SETR) \cite{zheng2021setr}, pioneering the use of a Transformer-based architecture for scene parsing, adopts a method similar to the vision Transformer (ViT) \cite{dosovitskiy2020vit} by tokenizing images into patches and processing them through Transformer blocks to enhance the global context modeling in the encoder. Furthermore, the MaskFormer series \cite{cheng2021maskformer, cheng2022mask2former} introduces a novel Transformer-based decoding paradigm by segmenting images into a set of masks, each associated with a class prediction. This mask classification paradigm, previously validated in \cite{li2023roadformer}, has been effectively incorporated into our enhanced RoadFormer+ design, further optimizing its performance.

\subsection{Data-Fusion Scene Parsing}

Scene parsing networks that rely solely on RGB images have been found to be highly sensitive to environmental factors such as lighting and weather conditions \cite{fan2020sneroadseg}. To overcome this limitation, data-fusion networks effectively utilize heterogeneous features extracted from RGB images and additional data sources. FuseNet \cite{hazirbas2017fusenet} pioneered the incorporation of depth information into scene parsing. It uses independent CNN encoders for RGB and depth images and fuses their features through element-wise summation. MFNet \cite{ha2017mfnet} and RTFNet \cite{sun2019rtfnet} strike a balance between speed and accuracy in RGB-Thermal driving scene parsing. Inspired by \cite{hazirbas2017fusenet}, the SNE-RoadSeg series \cite{fan2020sneroadseg,wang2021sneroadseg+,feng2024sne} incorporates surface normal information into freespace detection. These networks employ densely connected skip connections to enhance feature decoding. Despite the improved performance achieved by these networks, the simplistic feature fusion strategies potentially restrict their capacity to fully exploit the complementary information present in heterogeneous features.

To address this challenge, recent studies have employed more advanced and learnable feature fusion strategies. RoadFormer \cite{li2023roadformer} combines self-attention with channel attention to form a novel feature synergy block that greatly enhances the fusion of heterogeneous features. Data-fusion networks have also garnered attention in the broader domain of scene parsing. Recent works CMX \cite{zhang2023cmx} and CAINet \cite{lv2024context} utilize various attention modules to effectively fuse and recalibrate heterogeneous features. Additionally, SASEM \cite{yang2023exploring} introduces a plug-and-play module to enhance semantic supervision, thereby improving feature recovery capabilities in decoding. Moreover, CDDFuse \cite{zhao2023cddfuse} implements a two-step training strategy that integrates CNN and Transformer blocks in parallel to fuse multi-modal medical images effectively. This article delves into more robust and general-purpose modules so as to more effectively fuse heterogeneous features. Our proposed RoadFormer+ not only broadens its applicability and generalizability to a wider range of scene parsing tasks but also significantly reduces the number of model parameters.

\section{METHODOLOGY}
\label{Sect.methodology}
\begin{figure*}[t!]
\includegraphics[width=0.999\textwidth]{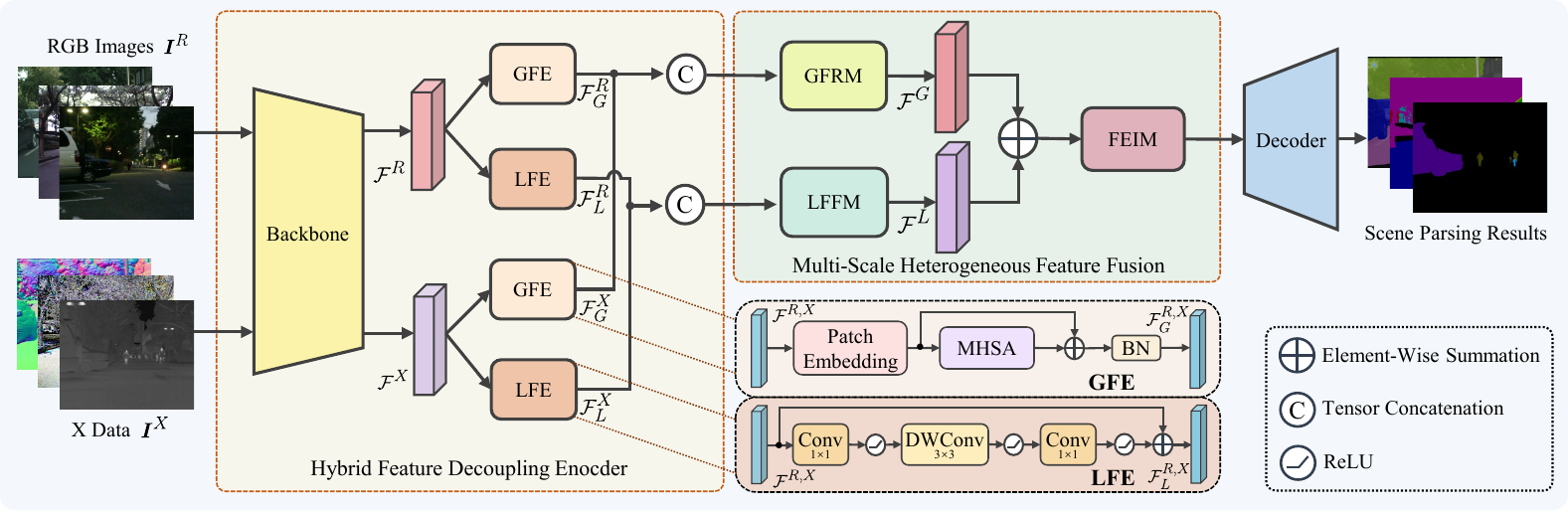}
\caption{An overview of our proposed RoadFormer+ architecture.}
\label{fig.structure}
\end{figure*}

As shown in Fig. \ref{fig.structure}, our proposed RoadFormer+ is composed of three key components:
\begin{itemize}
    \item An HFDE, which utilizes a weight-sharing backbone to extract heterogeneous features from RGB-X data and decouple them into global and local features using GFEs and LFEs;
    \item An MHFF block, which employs Transformer-based and CNN-based modules in parallel to recalibrate and enhance global and local features, respectively, and integrates them to generate fused features;
    \item A Transformer-based decoder that utilizes the fused features to generate final semantic predictions. Given the focus of this study on enhancing feature encoding and fusion strategies, we retain the decoder from the original RoadFormer \cite{li2023roadformer}.
\end{itemize}

\subsection{Hybrid Feature Decoupling Encoder}

\subsubsection{Overall Feature Encoding Pipeline}

Current networks generally employ symmetric duplex encoders to extract heterogeneous features from multiple data sources \cite{fan2020sneroadseg,li2023roadformer}. However, such dual-branch designs not only double the number of learnable parameters in the feature encoding phase but may also potentially lead to feature conflict \cite{wang2022rgb}. To address this issue, we propose an HFDE to improve the efficiency of heterogeneous feature extraction. Specifically, considering the correlation between heterogeneous features \cite{zhou2023cacfnet}, we first employ a weight-sharing backbone to process the given RGB image $\boldsymbol{I}^{R}\in\mathbb{R}^{H\times W \times 3}$ and its corresponding X data $\boldsymbol{I}^{X}\in\mathbb{R}^{H\times W \times 3}$, thereby generating multi-scale features $\mathcal{F}^{R}=\{ \boldsymbol{F}^R_1, \dots,\boldsymbol{F}^R_4 \}$ and $\mathcal{F}^{X} = \{ \boldsymbol{F}^X_1, \dots, \boldsymbol{F}^X_4 \}$. For X data with a single channel (such as depth, thermal, and polarization information), we replicate it three times along the channel dimension to match the RGB image's dimensions of $H\times W \times 3$. Here, $\boldsymbol{F}^{R, X}_i \in \mathbb{R}^{\frac{H}{S_i} \times \frac{W}{S_i} \times C_{i}}$ represents the features in the $i$-th encoding stage, where $C_{i}$ and $S_i=2^{i+1}$ ($i\in[1,4]\cap\mathbb{Z}$) denote the channel and stride numbers, respectively, and $H$ and $W$ denote the height and width of the input image, respectively. Furthermore, we employ two weight-separating GFEs and LFEs to extract global features $\mathcal{F}_{G}^{R,X}$ and local features $\mathcal{F}_{L}^{R,X}$ at four spatial scales from the heterogeneous features $\mathcal{F}^{R,X}$, respectively, thereby realizing feature decoupling. Finally, $\mathcal{F}_{G}^{R,X}$ and $\mathcal{F}_{L}^{R,X}$ are fed into the MHFF block for further feature fusion and recalibration.

\subsubsection{Weight-Sharing Backbone} 
Large-kernel convolutions exhibit considerable potential in capturing long-range dependencies, owing to their expansive receptive fields, while still retaining favorable inductive biases crucial for vision-specific tasks such as scene parsing \cite{ding2023unireplknet}. For instance, the areas surrounding vehicles are more likely to be roads rather than buildings. In our previous study \cite{li2023roadformer}, ConvNeXt \cite{liu2022convnet} demonstrates superior performance compared to ResNet \cite{he2016resnet} and Swin Transformer \cite{liu2021swin}, and thus, we continue to adopt it as the backbone in this study. We also compare the performance of ConvNeXt with the recently proposed SoTA backbone networks UniRepLKNet \cite{ding2023unireplknet} and DiNAT \cite{hassani2022dinat}. Detailed experimental results and analyses are provided in Table \ref{tab.HFDE}. Our backbone is constructed using two identical, weight-sharing ConvNeXt models.

\subsubsection{Global Feature Enhancer} 
ViT has shown exceptional performance across various fundamental vision tasks \cite{dosovitskiy2020vit, liu2021swin}. Its self-attention mechanism effectively models the global receptive field, thereby enhancing the contextual understanding essential for recognizing large continuous areas such as roads and sidewalks. Consequently, we utilize a GFE based on the multi-head self-attention mechanism to further emphasize the long-range global features. Given the robustness of the backbone network, we omit the positional encoding and replace the commonly used feed-forward network layers with simple normalization operations to reduce the number of model parameters. RGB features $\mathcal{F}^{R}$ and X features $\mathcal{F}^{X}$ are respectively mapped to query, value, and key matrices through convolutional layers. We also introduce a residual connection into the attention operation. Our GFE module can be formulated as follows:
\begin{equation}
\boldsymbol{G}_i  = \mathrm{Norm} \big( \mathrm{MHSA}
( \boldsymbol{F}_i )  + \boldsymbol{F}_i \big),
\end{equation}
where $\boldsymbol{F}_i$ represents the $i$-th feature maps within $\mathcal{F}^{R}$ and $\mathcal{F}^{X}$, $\boldsymbol{G}_i$ represents the $i$-th feature maps within $\mathcal{F}^{R}_G$ and $\mathcal{F}^{X}_G$, and $\mathrm{MHSA}$ represents the multi-head self-attention mechanism operation. After processing by the GFE, the enhanced global features $\mathcal{F}^{R}_G$ and $\mathcal{F}^{X}_G$ are obtained.

\subsubsection{Local Feature Extractor} 
Local detail features, such as edges and corners, are crucial for accurate scene parsing. Compared to Transformers, convolution operations are proficient at extracting local features and further enhancing fine-grained details \cite{Wu2020LiteTransformer}. Therefore, we propose a LFE, which incorporates an inverted residual block from MobileNetV2 \cite{sandler2018mobilenetv2} to process $\mathcal{F}^{R}$ and $\mathcal{F}^{X}$, specifically targeting local features. This lightweight module strikes a balance between model parameters and accuracy, as demonstrated across multiple tasks \cite{lv2024context}. Our LFE can be formalized as follows:
\begin{equation}
    \boldsymbol{L}_i
    = {\underset{1\times 1}{\mathrm{Conv}}}
    \bigg( \mathrm{ReLU}
    \Big( \underset{3\times 3}{\mathrm{DWConv}}
    \big( \mathrm{ReLU}
    (
    {\underset{1\times 1}{\mathrm{Conv}}}
    (\boldsymbol{F}_i)
    )
    \big)
    \Big) 
    \bigg) + \boldsymbol{F}_i,
\end{equation}
where $\boldsymbol{L}_i$ denotes the $i$-th feature maps within $\mathcal{F}^{R}_L$ and $\mathcal{F}^{X}_L$. After processing by the LFE, we obtain the local features $\mathcal{F}^{R}_L$ and $\mathcal{F}^{X}_L$.

\subsection{Multi-Scale Heterogeneous Feature Fusion Block}

To further fuse and integrate global and local features, we introduce a dual-branch MHFF block, which employs attention mechanisms and CNN modules in parallel. An MHFF consists of (1) a global feature recalibration module (GFRM) that utilizes a cross-attention mechanism to recalibrate $\mathcal{F}^{R}_G$ and $\mathcal{F}^{X}_G$, (2) a local feature fusion module (LFFM) that utilizes convolutional layers to fuse $\mathcal{F}^{R}_L$ and $\mathcal{F}^{X}_L$, and (3) a feature enhancement and integration module (FEIM) based on a spatial attention mechanism to integrate heterogeneous features, and generate robust fused feature $\mathcal{F}^{F}$.

\begin{figure*}[t!]
\includegraphics[width=0.99\textwidth]{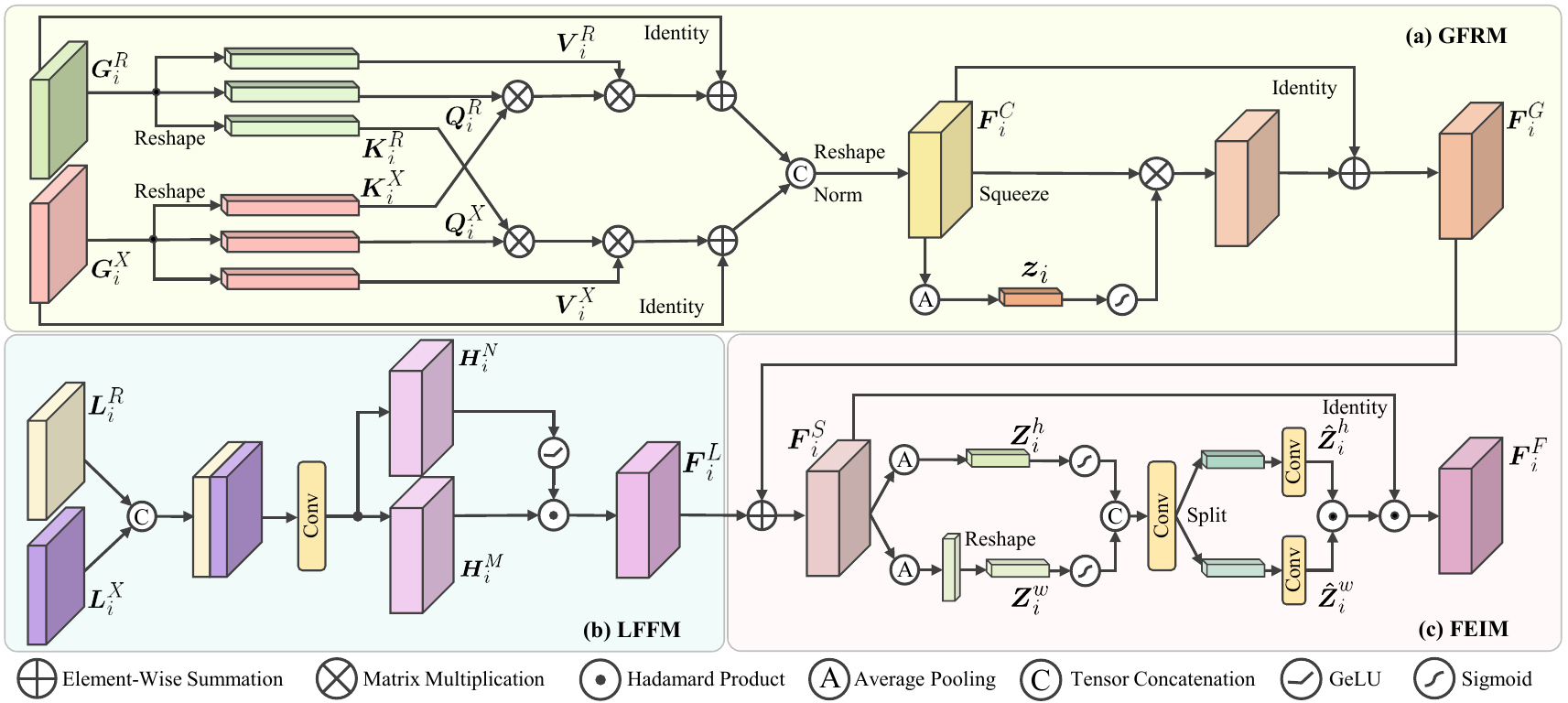}
\caption{An illustration of our proposed multi-scale heterogeneous feature fusion block, consisting of (a) a global feature recalibration module, (b) a local feature fusion module, and (c) a feature enhancement and integration module.}
\label{fig.structure_SHFF}
\end{figure*}

\subsubsection{Global Feature Recalibration Module}
Heterogeneous global features $\mathcal{F}^{R}_G$ and $\mathcal{F}^{X}_G$ are generally complementary \cite{zhou2023cacfnet}. For example, road areas often appear consistent in color across RGB images and possess uniform normal vectors and polarization properties. Therefore, one feature type can be utilized to mitigate potential noise in its complementary feature type \cite{zhang2023cmx}. Additionally, not all multi-channel features contribute positively to semantic predictions \cite{chen2020bi, li2023roadformer}, necessitating the recalibration of heterogeneous features along the channel dimension \cite{hu2018squeeze}. To address these challenges, we introduce the GFRM (see Fig. \ref{fig.structure_SHFF} (a)) to recalibrate and fuse $\mathcal{F}^{R}_G$ and $\mathcal{F}^{X}_G$. The cross-attention mechanism, which considers interactions among all positions in the input, is well-suited for calibrating complementary heterogeneous global features and has demonstrated excellent performance in many visual tasks \cite{zhang2023cmx}. Drawing inspiration from these approaches, the GFRM first recalibrates global features using a cross-attention mechanism, which can be formulated as follows:
\begin{equation}
\boldsymbol{G}^{R'}_i  = \mathrm{Softmax}
\left( \boldsymbol{Q}^R_i {\boldsymbol{K}^X_i}^\top \right) \kappa_i \boldsymbol{V}^R_i + \boldsymbol{G}^{R}_i,
\end{equation}
\begin{equation}
\boldsymbol{G}^{X'}_i  = \mathrm{Softmax}
\left( \boldsymbol{Q}^X_i {\boldsymbol{K}^R_i}^\top \right) \gamma_i \boldsymbol{V}^X_i + \boldsymbol{G}^{X}_i,
\end{equation}
\begin{equation}
\boldsymbol{F}^C_i = \mathrm{Norm} \left( \delta \big(  [\boldsymbol{G}^{R'}_i , \boldsymbol{G}^{X'}_i ]  \big) \right),
\end{equation}
where $\boldsymbol{G}^{R}_i$ and $\boldsymbol{G}^{X}_i$ represent the $i$-th feature maps within $\mathcal{F}^{R}_G$ and $\mathcal{F}^{X}_G$, respectively, $\boldsymbol{G}^{R}_i$ and $\boldsymbol{G}^{X}_i$ are then identically mapped to query $\boldsymbol{Q}^{R,X}_i$, key $\boldsymbol{K}^{R,X}_i$ and value $\boldsymbol{V}^{R,X}_i$ embeddings, $[\cdot,\cdot]$ denotes the concatenation operation along the channel dimension, and $\delta$ is a non-linear activation function. The learnable coefficients $\kappa_i$ and $\gamma_i$ can adaptively adjust attention significance \cite{fu2019dual}. For $\mathcal{F}^{C}$, we further employ channel-wise attention to emphasize key features and suppress those with low information density, which can be formulated as follows:
\begin{equation}
    {z}_{i,j}= \frac{S_i^2}{H W} \sum_{h=1}^{\frac{H}{S_i}}\sum_{w=1}^{\frac{W}{S_i}}\boldsymbol{F}^C_{i}(h,w,j),
    \label{eq:dcwfr4}
\end{equation}
\begin{equation}
    \boldsymbol{{F}}^G_i 
    = \boldsymbol{F}^C_i 
    \odot 
    \sigma
    \big(
    {\underset{1\times 1}{\mathrm{Conv}}}
    (\boldsymbol{z}_i)
    \big)
    + \boldsymbol{F}^C_i,
\end{equation}
where $\boldsymbol{z}_i=[z_{i,1},\dots,z_{i,C_i}]\in\mathbb{R}^{1\times 1\times C_i}$ stores the average pooling results of each feature map in $\boldsymbol{F}^C_{i}$, $\sigma$ is the sigmoid function, and $\odot$ denotes the Hadamard product operation. Finally, we obtain the fused global feature $\mathcal{F}^{G}=\{ \boldsymbol{F}^G_1, \dots,\boldsymbol{F}^G_4 \}$. Here, $\boldsymbol{F}^{G}_i \in \mathbb{R}^{\frac{H}{S_i} \times \frac{W}{S_i} \times C_{i}}$ represents the global features in the $i$-th feature fusion stages.

\subsubsection{Local Feature Fusion Module}
To preserve more local contexts when fusing heterogeneous features $\mathcal{F}^{R}_L$ and $\mathcal{F}^{X}_L$, we propose a convolution-based LFFM (see Fig. \ref{fig.structure_SHFF} (b)). Inspired by the MLP-Mixer \cite{tolstikhin2021mlpmixer}, our LFFM captures relationships between heterogeneous features from different local regions, generating fused local features. The LFFM can be mathematically represented as follows:
\begin{equation}
    \boldsymbol{H}^{L}_i =  \underset{3\times 3}{\mathrm{DWConv}}
    \big( 
    \mathrm{Conv}([\boldsymbol{L}^{R}_i , \boldsymbol{L}^{X}_i])
    \big),
\end{equation}
where $\boldsymbol{L}^{R}_i$ and $\boldsymbol{L}^{X}_i$ represent the $i$-th feature maps within $\mathcal{F}^{R}_L$ and $\mathcal{F}^{X}_L$, respectively, each having $C_i$ channels. After concatenating $\boldsymbol{L}^{R}_i$ and $\boldsymbol{L}^{X}_i$ along the channel dimension, we employ depth-wise separable convolutions to expand their channels to $4C_i$, thereby enhancing the local context. The resultant $\boldsymbol{H}^{L}_i$ is then split into $\boldsymbol{H}^{M}_i$ and $\boldsymbol{H}^{N}_i$ along the channel dimension. This design allows the model to learn new feature representations, the effectiveness of which is further validated in Table \ref{tab.SFFM_ablation_3}. These two components interact through Hadamard multiplication, enabling the capture of relationships between features from different local regions:
\begin{equation}
    \boldsymbol{F}^{L}_i =  {\underset{1\times 1}{\mathrm{Conv}}}
    \big(
    \boldsymbol{H}^{M}_i \odot \sigma(\boldsymbol{H}^{N}_i)
    \big),
\end{equation}
where we utilize the Gaussian error linear unit (GELU) as the activation function $\sigma(\cdot)$. Then we obtain the fused local features $\mathcal{F}^{L}=\{ \boldsymbol{F}^L_1, \dots,\boldsymbol{F}^L_4 \}$, where $\boldsymbol{F}^{L}_i \in \mathbb{R}^{\frac{H}{S_i} \times \frac{W}{S_i} \times C_{i}}$ represents the local features in the $i$-th feature fusion stage.

\begin{figure*}[t!]
\includegraphics[width=0.999\textwidth]{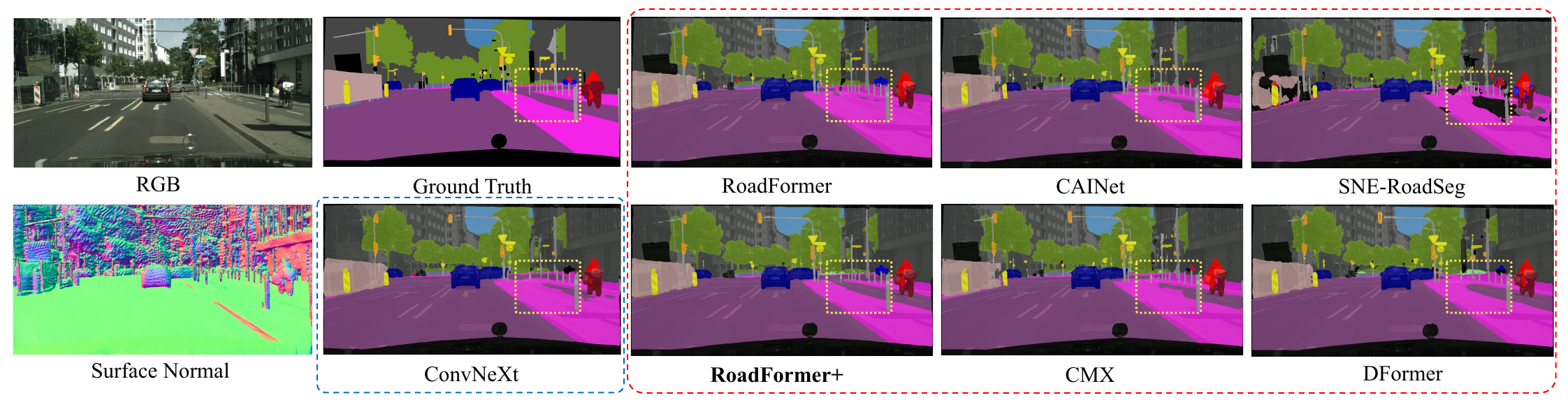}
\caption{Qualitative comparisons between our proposed RoadFormer+ and other SoTA networks on the Cityscapes validation set, where significantly improved regions are shown with yellow dashed boxes.}
\label{fig.visualizationcityscapes}
\end{figure*}
\subsubsection{Feature Enhancement and Integration Module}
Spatial information is crucial for capturing spatial structures in visual perception tasks \cite{li2023ccafusion}. Nonetheless, our GFRM and LFFM fuse heterogeneous features across channel dimensions, squeezing spatial information into a channel descriptor, and hence is difficult to preserve positional information \cite{hou2021coordinate}. Therefore, it is necessary to introduce additional spatial information when integrating global features $\mathcal{F}^{G}$ and local features $\mathcal{F}^{L}$. The spatial attention mechanism emphasizes the importance of specific regions within features, aiding the network in focusing on ``where'' informative parts are located \cite{woo2018cbam}. Inspired by the coordinated attention \cite{hou2021coordinate}, we introduce the FEIM (see Fig. \ref{fig.structure_SHFF} (c)) to further enhance and integrate $\mathcal{F}^{G}$ and $\mathcal{F}^{L}$, enabling the module to detect more subtle spatial variations. Specifically, we employ global pooling kernels $(H, 1)$ or $(1, W)$ to aggregate features along the height and width dimensions, respectively. Thus, the output of the $j$-th channel at height $p$ and width $q$ can be formulated as:
\begin{equation} 
\boldsymbol{F}^{S}_i = \boldsymbol{F}^{G}_i + \boldsymbol{F}^{L}_i,
\end{equation}
\begin{equation} 
z_{i,j,p}^h = \frac{1}{W} \sum_{0 \le m < W}\boldsymbol{F}^S_i(p, m, j),
\end{equation}
\begin{equation} 
z_{i,j,q}^w = \frac{1}{H} \sum_{0 \le n < H}\boldsymbol{F}^S_i(n, q, j),
\end{equation}
where $\boldsymbol{F}^{G}_i$ and $\boldsymbol{F}^{L}_i$ represent the $i$-th feature maps within $\mathcal{F}^{G}$ and $\mathcal{F}^{L}$, respectively, and $\boldsymbol{Z}^h_i\in\mathbb{R}^{H\times 1\times C_i}$ as well as $\boldsymbol{Z}^w_i\in\mathbb{R}^{1 \times W \times C_i}$ store the average pooling results of each feature map in $\boldsymbol{F}^S_{i}$ across the dimensions of $H$ and $W$, respectively. $\boldsymbol{Z}^w_i$ is subsequently reshaped into $\mathbb{R}^{W \times 1 \times C_i}$. We further apply a convolutional layer and a Sigmoid function to make full use of the captured positional information, enhancing the network's ability to accurately emphasize regions of interest. This process can be formulated as follows:
\begin{equation}
\boldsymbol{\hat{Z}}_i = \sigma \big( \underset{1\times 1}{\mathrm{Conv}} ( [\boldsymbol{Z}^h_i, \boldsymbol{Z}^w_i] ) \big),
\end{equation}
where $[\cdot,\cdot]$ denotes the concatenation operation along the spatial dimension. Then, $\boldsymbol{\hat{Z}}_i$ is split into two separate tensors: $\boldsymbol{\hat{Z}}^{h}_i\in\mathbb{R}^{H\times 1\times C_i}$ and $\boldsymbol{\hat{Z}}^{w}_i\in\mathbb{R}^{W\times 1\times C_i}$. This allows interactions between $\boldsymbol{\hat{Z}}^{h}_i$ and $\boldsymbol{\hat{Z}}^{w}_i$ from both dimensions, enhancing the emphasis on regions of interest. $\boldsymbol{\hat{Z}}^{w}_i$ is then reshaped into $\mathbb{R}^{1\times W\times C_i}$. Each element within the two attention maps, $\boldsymbol{\hat{Z}}^{h}_i$ and $\boldsymbol{\hat{Z}}^{w}_i$, indicates the presence of objects of interest across respective rows or columns. $\boldsymbol{\hat{Z}}^{h}_i$ and $\boldsymbol{\hat{Z}}^{w}_i$ are applied to $\boldsymbol{F}^S_i$ to more accurately pinpoint the exact location of the object of interest, which can be written as follows:
\begin{equation} \label{eqn:attention}
\boldsymbol{F}^F_i = \boldsymbol{F}^S_i \odot \boldsymbol{\hat{Z}}^{h}_i \odot \boldsymbol{\hat{Z}}^{w}_i.
\end{equation}
Finally, we obtain the fused features $\mathcal{F}^{F}=\{ \boldsymbol{F}^F_1, \dots,\boldsymbol{F}^F_4 \}$, which are forwarded to the decoder to obtain the final semantic prediction. Given the outstanding performance of the multi-scale Transformer decoder employed in RoadFormer, we retain this design. Please refer to \cite{li2023roadformer} for more details on the decoder and loss function.

\section{EXPERIMENTS}
\label{Sect.experments}
\subsection{Datasets}
\label{Sect.dataset}

We first compare RoadFormer+ with other SoTA scene parsing networks on the following seven RGB-X datasets:
\subsubsection{SYN-UDTIRI \cite{li2023roadformer}}
\label{Sect. synudtiri}

This dataset contains over 10,000 pairs of stereo road images, along with corresponding depth maps, surface normal information, and semantic annotations, including three categories: freespace, road defect, and other objects. It is created using the CARLA simulator \cite{dosovitskiy2017carla} and first introduced in our previous work \cite{li2023roadformer}. The input images are resized to a resolution of 640$\times$352 pixels.

\subsubsection{KITTI Road \cite{geiger2013kitti}}
\label{Sect. kitti}

This dataset has 289 pairs of stereo road images and their corresponding LiDAR point clouds for both model training and validation. We employ a data preprocessing strategy akin to that detailed in \cite{fan2020sneroadseg}. The input images are resized to a resolution of 1,280$\times$384 pixels.

\subsubsection{Cityscapes \cite{cordts2016cityscapes}}
\label{Sect. cityscapes_dataset}

This widely used urban scene dataset contains 2,975 stereo training images and 500 validation images, with well-annotated semantic annotations. Notably, the surface normal information is derived from depth images generated using RAFT-Stereo \cite{lipson2021raft}, trained on the KITTI dataset \cite{menze2015object}. The input images are resized to a resolution of 1,024$\times$512 pixels.

\subsubsection{KITTI Semantics \cite{abu2018augmented}}
\label{Sect. kittisemantics}
This dataset contains 200 images and their corresponding semantic annotations across 19 classes. Surface normal information is derived from depth data acquired using ViTAStereo \cite{liu2024playing}, chosen for its superior accuracy. The input images are resized to a resolution of 1,280$\times$384 pixels.

\subsubsection{MFNet \cite{ha2017mfnet}}
\label{Sect. mfnet}

This urban driving scene parsing dataset contains 1,569 synchronized pairs of RGB and thermal images at a resolution of 640$\times$480 pixels. It includes semantic annotations across nine classes: bike, person, car, road lanes, guardrail, car stop, bump, color cone, and background.

\subsubsection{FMB \cite{liu2023multi}} 
\label{Sect. fmb}

This dataset contains 1,500 well-rectified RGB-Thermal image pairs (resolution: 800$\times$600 pixels), captured in urban driving scenes under different illumination conditions. It provides semantic annotations across 14 classes.

\subsubsection{ZJU \cite{xiang2021polarization}} 
\label{Sect. zju}
This RGB-Polarization dataset, designed for automated driving applications, containing 394 image pairs. Each pair contains four polarized images captured at different polarization angles ($0^\circ$, $45^\circ$, $90^\circ$, and $135^\circ$). We use the angle of linear polarization (AoLP) algorithm to obtain the polarization information. The input images are resized to a resolution of 612$\times$512 pixels.

\subsection{Experimental Setup and Evaluation Metrics}
\label{Sect.setup}

For the SYN-UDTIRI and other RGB-Normal datasets, we exclusively use surface normal information estimated using the D2NT algorithm \cite{icra_2023_D2NT} owing to its superior accuracy. This information serves as the ``X'' data to train our RoadFormer+. Additionally, depth, thermal, and polarization information are replicated across the channel dimension three times during data preprocessing to match the $H \times W \times 3$ dimensions of RGB images. During training, we employ the common data augmentation techniques used in semantic segmentation, including resizing, random cropping, and random flipping of RGB-X image pairs. Additionally, we make random adjustments to the brightness, contrast, saturation, and hue of the RGB images. All networks are trained for the same number of epochs on an NVIDIA RTX 3090 GPU using the AdamW optimizer \cite{loshchilov2018decoupled}, with a polynomial decay strategy for the learning rate \cite{chen2017deeplab}. The initial learning rate is set at $10^{-4}$ with a weight decay of $5 \times 10^{-2}$, and learning rate multipliers of $10^{-1}$ are applied to the weight-sharing backbone.

We evaluate the performance of our models using five common metrics: accuracy (Acc), precision (Pre), recall (Rec), intersection over union (IoU), and F-score (Fsc). We refer readers to our previous work \cite{li2023roadformer} for more details on these metrics. Additionally, the evaluation metrics used for the KITTI Road and KITTI Semantics benchmarks are available on the official webpage: \url{cvlibs.net/datasets/kitti}.

\begin{table*}[t!]
\fontsize{7.5}{11}\selectfont
\centering
\caption{Quantitative comparison of \textbf{road defect detection} on the SYN-UDTIRI test set.}
\begin{tabular}{C{2.8cm}|C{1.8cm}|C{1.4cm}C{1.4cm}C{1.4cm}C{1.4cm}|C{1.6cm}}
\toprule[1pt]
 Method  &  Publication  & IoU (\%) $\uparrow$ & Fsc (\%) $\uparrow$ & Pre (\%) $\uparrow$ & Rec (\%)  $\uparrow$ & \#Params (M) $\downarrow$  \\

\cline{1-7}
 FuseNet & ACCV'16 \cite{hazirbas2017fusenet} & 70.70 & 82.90 & 72.10  & \textbf{97.50} & 50.0   \\
 OFF-Net & ICRA'22 \cite{min2022orfd} & 83.80 & 91.20 & 91.90 & 90.50 & 25.2  \\
 MFNet & IROS'17 \cite{ha2017mfnet} & 87.70  & 93.50 & 96.20 & 90.90 & \textbf{0.8}   \\
 RTFNet & RAL'19 \cite{sun2019rtfnet} & 90.50 & 95.00 & 95.50 & 94.50 & 254.5   \\
 DFormer & ICLR'24 \cite{yin2024dformer} & 90.88 & 95.22  & 96.09 & 94.37 & 38.8 \\
 CAINet & T-MM'24 \cite{lv2024context} & 91.77 & 95.71 & 95.43 & 95.99 & 12.2 \\
 SNE-RoadSeg & ECCV'20 \cite{fan2020sneroadseg} & 92.10 & 95.90 & 96.70 & 95.10 &  201.3  \\
 CMX & T-ITS'23 \cite{zhang2023cmx} & 93.31 & 96.27 & 96.54 & 96.81 & 138.8   \\
 RoadFormer (B)  & T-IV'24 \cite{li2023roadformer} & 93.06 & 96.41 & 96.19 & 96.63  & 206.8 \\
 RoadFormer (L)  & T-IV'24 \cite{li2023roadformer} & 93.51 & 96.65 & 96.61 & 96.69  & 438.6  \\
\cline{1-7}
\rowcolor{gray!20}  \textbf{RoadFormer+ (B)}  & Ours & \textbf{94.11} & \textbf{96.96} & \textbf{97.03} & 96.90  & 152.4 \\
\bottomrule[1pt]
\end{tabular}
\label{tab.synudtiri}
\end{table*}

\begin{table}[t!]
\fontsize{7.5}{11}\selectfont
\centering
\caption{Comparison with SoTA algorithms published on the KITTI Road benchmark.}
{
\setlength{\tabcolsep}{6pt}
\begin{tabular}{l|ccc|c}
\toprule[1pt]
Method & MaxF (\%) $\uparrow$ & Pre (\%) $\uparrow$ & Rec (\%) $\uparrow$ & Rank \\ 
\hline
NIM-RTFNet \cite{wang2020applying}  & 96.02  & 96.43 & 95.62 & 19 \\ 
SNE-RoadSeg \cite{fan2020sneroadseg}  & 96.75  & 96.90 & 96.61 & 13 \\ 
LRDNet+ \cite{khan2022lrdnet}  & 96.95 & 96.88 & 97.02 & 9 \\
PLB-RD \cite{sun2022pseudo}  & 97.42 & 97.30 & 97.54 & 5 \\
RoadFormer (B) \cite{li2023roadformer}  & 97.50 & 97.16 & \textbf{97.84} & 3 \\
SNE-RoadSegV2 \cite{feng2024sne}  & 97.55 & \textbf{97.57} & 97.53 & 2 \\
\hline
\rowcolor{gray!20} \textbf{RoadFormer+ (B)} & \textbf{97.56} & 97.43 & 97.69 & {\textbf{1}} \\
\bottomrule[1pt]
\end{tabular}
}
\label{tab.kitti}
\end{table}

\begin{table}[t!]
\fontsize{7.5}{11}\selectfont
\centering
\caption{Quantitative comparison of \textbf{freespace detection} on the validation set of the Cityscapes dataset. 
}
{
\setlength{\tabcolsep}{6pt}
\begin{tabular}{
L{2.5cm}|ccc
}
\toprule[1pt]
Method  & IoU (\%) $\uparrow$& Fsc (\%) $\uparrow$& Acc  (\%) $\uparrow$\\
\hline
NIM-RTFNet \cite{wang2020applying} & 91.43 & 92.02 & 96.07 \\
LRDNet+ \cite{khan2022lrdnet} & 92.82 & 94.71 & 97.02   \\
PLB-RD \cite{sun2022pseudo} & 92.96 & 95.28 & 97.15  \\
SNE-RoadSeg \cite{fan2020sneroadseg} & 93.22 & 96.49 & 97.68   \\
SNE-RoadSegV2 \cite{feng2024sne} & 94.40 & 97.12 & 98.11   \\
RoadFormer (B) \cite{li2023roadformer}  & 95.87 & 97.89 & \textbf{98.30}  \\
\hline
\rowcolor{gray!20} \textbf{RoadFormer+ (B)} & \textbf{96.01} & \textbf{97.96} & 97.82  \\
\bottomrule[1pt]
\end{tabular}
}
\label{tab.cityscapesroads}
\end{table}

\begin{table}[t!]
\fontsize{7.5}{11}\selectfont
\centering
\caption{Quantitative comparison of \textbf{all-category scene parsing} on the validation set of the Cityscapes dataset. 
}
{
\setlength{\tabcolsep}{5pt}
\begin{tabular}{
l|l|ccc
}
\toprule[1pt]
& Method  & mIoU (\%) $\uparrow$ & mFsc (\%) $\uparrow$ & mAcc (\%) $\uparrow$ \\
\hline
\multirow{4}{*}{\rotatebox[origin=c]{90}{RGB}}
& SegFormer \cite{xie2021segformer} & 64.51 & 76.99 & 76.39 \\
& DeepLabV3+ \cite{chen2017deeplab} & 68.66 & 80.34  & 78.89 \\
& ConvNeXt \cite{liu2022convnet} & 73.35 & 83.94 & 83.32 \\
& Mask2Former \cite{cheng2022mask2former} & 74.78 & 84.97 & 85.90  \\
\hline
\multirow{5}{*}{\rotatebox[origin=c]{90}{RGB-Depth}} 
& CAINet \cite{lv2024context} & 62.38 & 75.04 & 73.68   \\
& CMX \cite{zhang2023cmx} & 74.11 & 84.41 & 83.30   \\
& DFormer \cite{yin2024dformer} & 74.37 & 84.55 & 84.00   \\
& RoadFormer (B) \cite{li2023roadformer} & 76.09 & 85.83 & 86.30 \\
\cline{2-5}
& \textbf{RoadFormer+ (B)} & 77.42 & 86.72 & 86.23  \\
\hline
\multirow{7}{*}{\rotatebox[origin=c]{90}{RGB-Normal}} 
& RTFNet \cite{sun2019rtfnet} & 49.60 & 61.20 & \textbf{90.00}   \\
& SNE-RoadSeg \cite{fan2020sneroadseg} & 53.40 & 64.54  & 85.64  \\
& CAINet \cite{lv2024context} & 62.41 & 75.13 & 74.23   \\
& CMX \cite{zhang2023cmx} & 73.50 & 83.99 & 83.67  \\
& RoadFormer (B) \cite{li2023roadformer} & 76.18 & 85.88 & 85.38  \\
\cline{2-5}
& \textbf{RoadFormer+ (B)} & {77.57} & {86.84}  & {86.77} \\
& \cellcolor{gray!20}\textbf{RoadFormer+ (L)} & \cellcolor{gray!20}\textbf{78.53} &\cellcolor{gray!20}\textbf{87.48}  & \cellcolor{gray!20}{87.00}  \\
\bottomrule[1pt]
\end{tabular}
}
\label{tab.cityscapes}
\end{table}

\begin{table}[t!]
\fontsize{7.4}{11}\selectfont
\centering
\caption{Comparison with SoTA algorithms published on the KITTI Semantics benchmark.}
{
\setlength{\tabcolsep}{6pt}
\begin{tabular}{l|cc|c}
\toprule[1pt]
Method & IoU Class (\%) $\uparrow$ & IoU Category (\%) $\uparrow$ & Rank \\ 
\hline
SegStereo \cite{yang2018segstereo} & 59.10 & 81.31 & 11 \\ 
Chroma UDA \cite{erkent2020semantic} & 60.36 & 80.73 & 8 \\
MSeg \cite{lambert2020mseg} & 62.64 & 86.59 & 6 \\
RoadFormer (B) \cite{li2023roadformer} & 67.17 & 87.89 & 5 \\
SNp-DN161 \cite{bevandic2022multi} & 68.89 & 87.06 & 5 \\
VideoProp-LabelRelax \cite{zhu2019improving} & 72.82 & \textbf{88.99} & 4 \\
\hline
 \textbf{RoadFormer+ (B)} & 70.32 & 87.16 & - \\
\rowcolor{gray!20} \textbf{RoadFormer+ (L)} & \textbf{73.13} & 88.75 & \textbf{3} \\
\bottomrule[1pt]
\end{tabular}
}
\label{tab.kittisemantics}
\end{table}

\begin{table}[t!]
\fontsize{7.5}{11}\selectfont
\centering
\caption{Quantitative comparison on the MFNet test set. 
}
{
\setlength{\tabcolsep}{5pt}
\begin{tabular}{
L{2.3cm}|c|c
}
\toprule[1pt]
Method  & mIoU (\%) $\uparrow$ & Rank  \\
\hline
RTFNet \cite{sun2019rtfnet} & 53.2 &  33 \\
MDBFNet \cite{liang2024multi} & 57.8 &  13 \\
RoadFormer (B) \cite{li2023roadformer} & 58.0 &  12 \\
CAINet \cite{lv2024context} & 58.6 &  9 \\
CMX \cite{zhang2023cmx} & 59.7 & 5  \\	
CMNeXt \cite{zhang2023delivering} & 59.9 & 4  \\
CRM-RGBTSeg \cite{shin2023complementary} & 61.4 & 3  \\
HAPNet \cite{li2024hapnet} & 61.5 & 2  \\
\hline
\textbf{RoadFormer+ (B)} & {60.9} & -  \\
\rowcolor{gray!20} \textbf{RoadFormer+ (L)} & \textbf{62.7} & \textbf{1}  \\
\bottomrule[1pt]
\end{tabular}
}
\label{tab.mfnet}
\end{table}

\begin{table}[t!]
\fontsize{7.5}{11}\selectfont
\centering
\caption{Quantitative comparison on the FMB dataset. 
}
{
\setlength{\tabcolsep}{5pt}
\begin{tabular}{
L{2.3cm}|c|c
}
\toprule[1pt]
Method  & mIoU (\%) $\uparrow$  & Rank \\
\hline
DIDFuse \cite{li2021didfuse} & 50.6 & 6 \\
ReCoNet \cite{huang2022reconet} & 50.9  & 5 \\
SegMiF \cite{liu2023multi} & 54.8 & 4 \\
MMSFormer \cite{reza2024mmsformer} & 61.7 & 3 \\
RoadFormer (B) \cite{li2023roadformer} & 69.2 & 2 \\
\hline
\textbf{RoadFormer+ (B)} & {73.1} & - \\
\rowcolor{gray!20} \textbf{RoadFormer+ (L)} & \textbf{74.1} & \textbf{1} \\
\bottomrule[1pt]
\end{tabular}
}
\label{tab.fmb}
\end{table}

\begin{table}[t!]
\fontsize{7.5}{11}\selectfont
\centering
\caption{Quantitative comparison on the ZJU-RGB-P dataset. 
}
{
\setlength{\tabcolsep}{5pt}
\begin{tabular}{
L{2.3cm}|c|c
}
\toprule[1pt]
Method  & mIoU (\%) $\uparrow$  & Rank \\
\hline
EAFNet \cite{xiang2021polarization} & 85.7 & 5 \\
RoadFormer (B) \cite{li2023roadformer} & 92.6 & 4 \\
CMX \cite{zhang2023cmx} & 92.6  & 3 \\
ShareCMP \cite{liu2023sharecmp} & 92.7 & 2 \\
\hline
\textbf{RoadFormer+ (B)} & {92.9} & - \\
\rowcolor{gray!20} \textbf{RoadFormer+ (L)} & \textbf{93.0} & \textbf{1}   \\
\bottomrule[1pt]
\end{tabular}
}
\label{tab.zju}
\end{table}

\subsection{Comparison with SoTA Networks}
We first conduct experiments on four RGB-Normal datasets. The quantitative results on the SYN-UDTIRI, Cityscapes, KITTI Road, and KITTI Semantics datasets are presented in Tables \ref{tab.synudtiri}-\ref{tab.kittisemantics}, respectively. In these experiments, the symbols ``B'' and ``L'' respectively denote the use of ConvNeXt-B and ConvNeXt-L as the backbones. These results demonstrate that our proposed RoadFormer+ significantly outperforms all other SoTA networks, including our previous work RoadFormer \cite{li2023roadformer}, across all four RGB-Normal datasets. This validates its exceptional performance and robustness in effectively parsing various types of road scenes. Notably, as shown in Tables \ref{tab.synudtiri}, RoadFormer+ based on ConvNeXt-B reduces the number of learnable parameters by 65\% compared to RoadFormer.

Furthermore, we conduct experiments on the Cityscapes dataset by treating it as both a binary segmentation task (road versus background) and a full-category segmentation task (19 labeled categories plus an ``ignore'' category). Experimental results are presented in Tables \ref{tab.cityscapes} and \ref{tab.cityscapesroads}, respectively. We also compare RoadFormer+ with four SoTA single-modal networks. It is worth noting that traditional data-fusion networks, which typically employ basic element-wise addition or feature-level concatenation for feature fusion, perform worse than single-modal networks. This underperformance may be attributed to the noise present in disparity maps used for surface normal estimation, which are derived directly from a stereo-matching network pre-trained on the KITTI dataset. Experimental results further demonstrate that RoadFormer+ effectively overcomes this issue through feature recalibration and enhancement, thus preventing performance degradation even when surface normal information is inaccurate.

We submit the test set results obtained by RoadFormer+ to both the KITTI Road and KITTI Semantics benchmarks for performance comparison. As shown in Tables \ref{tab.kitti} and \ref{tab.kittisemantics}, RoadFormer+ ranks first on the KITTI Road benchmark and ranks third on the KITTI Semantics benchmark. Notably, the top-performing SoTA methods in the KITTI Semantics benchmark employ sequential frames (±10) from the scene flow subset for data augmentation. Despite this, RoadFormer+ exhibits superior performance in urban scene parsing compared to all previously published methods.

Furthermore, we explore the applicability of RoadFormer+ for RGB-Thermal and RGB-Polarization scene parsing. Experimental results on three public datasets, MFNet (RGB-Thermal), FMB (RGB-Thermal), and ZJU (RGB-Polarization), demonstrate the superiority of RoadFormer+ over other task-specific data-fusion networks for these modalities. Impressively, RoadFormer+ achieves an improvement in mIoU of 1.2-9.5\% on the MFNet dataset, 3.9-22.5\% on the FMB dataset, and 0.3-7.3\% on the ZJU dataset, compared to other SoTA methods. These results underscore the versatility of our network in handling diverse data types. It is important to note that since the ``bicycle'' category is not included in the test set of the FMB dataset, we report the mIoU metrics excluding  the ``bicycle'' category.

\begin{figure}[t!]
\includegraphics[width=0.485\textwidth]{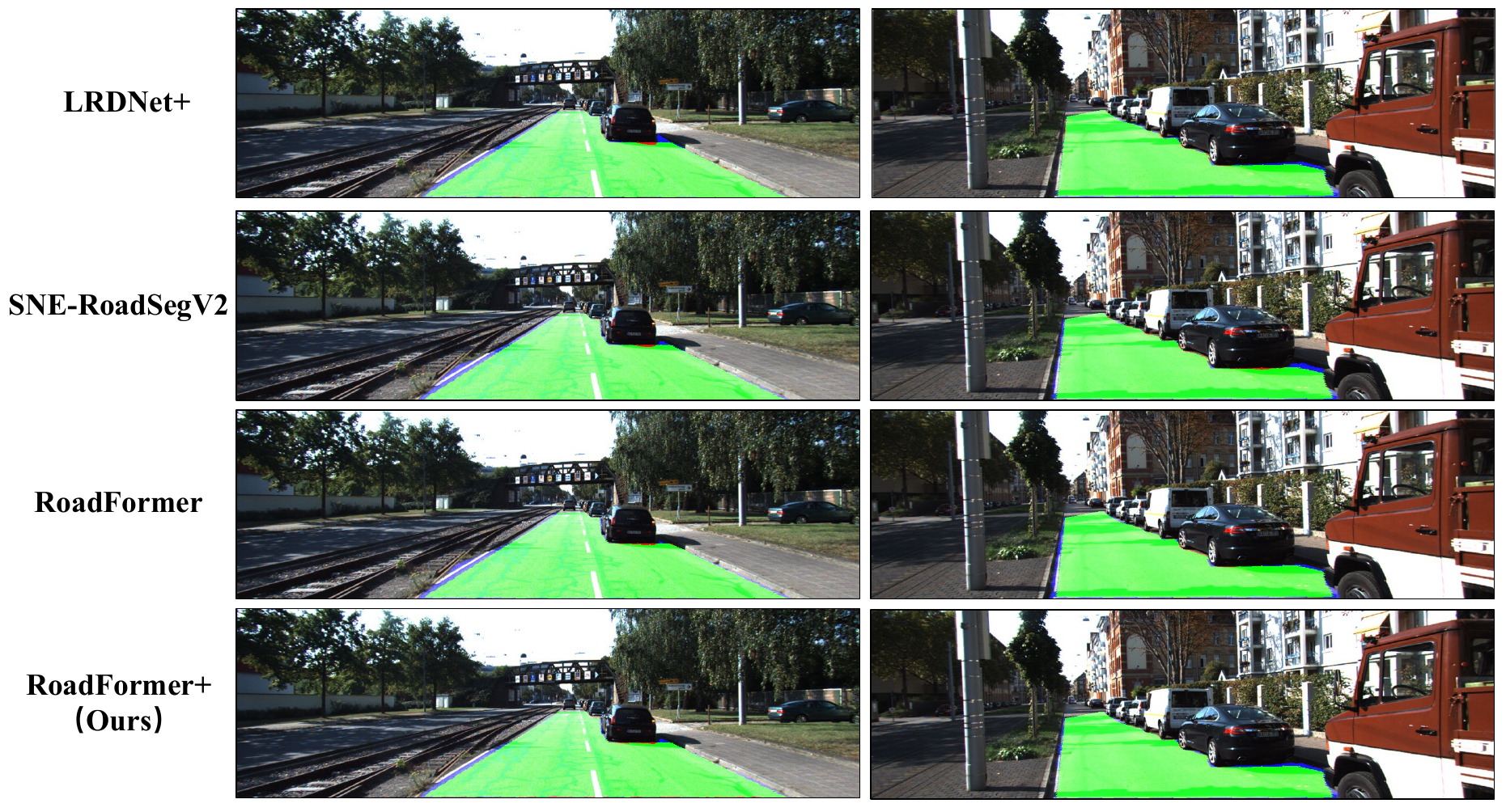}
\caption{Qualitative comparison between our proposed RoadFormer+ and other SoTA networks on the KITTI Road dataset. The results are produced by the official KITTI online benchmark suite. The classifications are visualized with true positives in green, false positives in blue, and false negatives in red.}
\label{fig.visualizationkitti}
\end{figure}

\begin{figure*}[t!]
\includegraphics[width=0.999\textwidth]{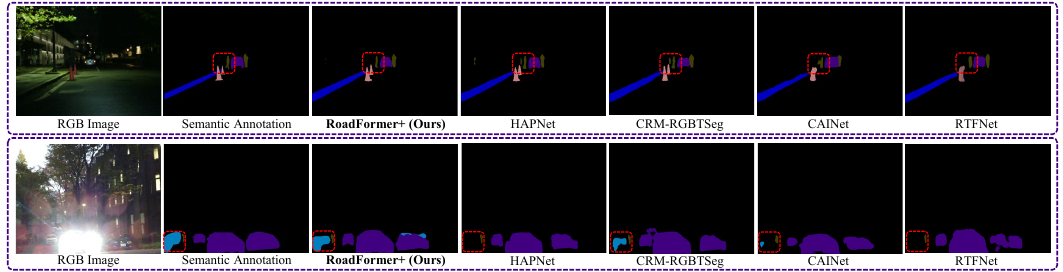}
\caption{Qualitative comparisons between our proposed RoadFormer+ and other SoTA networks on the MFNet test set, with significantly improved regions highlighted in red dashed boxes.}
\label{fig.visualizationmfnet}
\end{figure*}

Qualitative comparisons on the KITTI Road, Cityscapes, and MFNet datasets are presented in Figs. \ref{fig.visualizationkitti}-\ref{fig.visualizationmfnet}. The dual-branch feature fusion design of RoadFormer+ enables effective capture of both local and global contexts, thereby outperforming previous single-branch heterogeneous feature fusion approaches. Our method not only demonstrates robust performance in comprehensive scene understanding but also excels in delineating detailed boundaries. Additionally, RoadFormer+ exhibits superior capabilities in handling challenging conditions such as darkness and fog, demonstrating its versatility across diverse environmental scenarios. Furthermore, RoadFormer+ consistently delivers robust performance across various illumination conditions. As illustrated in the second row of Fig \ref{fig.visualizationmfnet}, RoadFormer+ outperforms all existing data-fusion methods in handling overexposed scenes within the MFNet dataset.

\subsection{Ablation Studies}
We conduct ablation studies on the SYN-UDTIRI, MFNet, and ZJU datasets. Our baseline is built upon RoadFormer \cite{li2023roadformer}, and all implementation details are consistent with those described in Sect. \ref{Sect.setup}.

\subsubsection{Effectiveness of HFDE}

Building on our previous findings stated in \cite{li2023roadformer} that demonstrated the effectiveness of ConvNeXt \cite{liu2022convnet} in urban scene parsing, we continue to employ it as the backbone in this study. We investigate two backbone training strategies: weight-sharing and weight-separating. The results, presented in Table \ref{tab.HFDE}, show that the weight-sharing strategy not only achieves performance comparable to the weight-separating strategy across three RGB-X datasets but also significantly reduces the model's parameters by nearly half. This observation calls into question the utility of traditional duplex encoder designs in these applications. 

We further validate the effectiveness of our proposed GFE and LFE on the SYN-UDTIRI dataset in terms of heterogeneous feature enhancement. It is evident that using either GFEs or LFEs independently can effectively enhance our model's performance, and their combined use results in an IoU increase of $0.57\%$. Additionally, we compare ConvNeXt with recently proposed models, including the Transformer-based DiNAT \cite{hassani2022dinat} and UniRepLKNet \cite{ding2023unireplknet}, which both employ large-kernel convolutions. The results affirm that ConvNeXt continues to exhibit superior performance compared to other backbones.

\begin{table}[t!]
\fontsize{7}{11}\selectfont
\centering
\caption{
Ablation study on the backbone training strategy when using ConvNeXt as the backbone.}
{
\setlength{\tabcolsep}{3pt} 
\begin{tabular}{c|c|c|c|c}
\toprule[1pt]
Strategy  & \makecell{SYN-UDTIRI \\ IoU (\%)$\uparrow$}  & \makecell{MFNet \\ mIoU (\%)$\uparrow$}  & \makecell{ZJU \\ mIoU (\%)$\uparrow$} & \#Params (M)* $\downarrow$   \\
\hline
Weight-Separating & \textbf{92.88} & {58.88} & \textbf{92.54} & 206.8  \\
Weight-Sharing & 92.87 & \textbf{58.96} & {92.47} & \textbf{113.7} \\
\bottomrule[1pt]
\end{tabular}
}
\\[5pt] 
\footnotesize{*The resolution of the input image is set to 640$\times$352 pixels.}
\label{tab.HFDE_ablation}
\end{table}

\begin{table}[t!]
\fontsize{7.5}{11}\selectfont
\centering
\caption{Ablation study on the backbone selection and the effectiveness of our proposed HFDE.}
\begin{tabular}
{M{2.2cm}|cc|c|c}
\toprule[1pt]
Backbone & GFE & LFE & IoU (\%) $\uparrow$  & \#Params (M) $\downarrow$   \\
\hline
ConvNeXt-B & $\checkmark$ & $\times$  & 93.05  & \textbf{123.8}  \\
ConvNeXt-B  & $\times$ & $\checkmark$  & 93.13   & 124.9 \\
ConvNeXt-B  & $\checkmark$ & $\checkmark$  & \textbf{93.44}   & 134.9  \\
\cline{1-5}
DiNAT-B & $\checkmark$ & $\checkmark$  & 93.19  & 136.1  \\
UniRepLKNet-B & $\checkmark$  & $\checkmark$  & 93.36  & 145.5  \\
\bottomrule[1pt]
\end{tabular}
\label{tab.HFDE}
\end{table}

\subsubsection{Effectiveness of the MHFF Block}

As illustrated in Table \ref{tab.SFFM_ablation_3}, we utilize RoadFormer as the baseline and alternately replace its feature fusion module with components from our proposed MHFF block to validate the efficacy of the dual-branch feature fusion design. First, we maintain RoadFormer's HFFM and FFRM to fuse global and local features, with the results depicted in the first row. As indicated in the second row, we maintain the use of the HFFM for global feature fusion while integrating the proposed LFFM for local feature fusion, resulting in performance improvements on the SYN-UDTIRI and MFNet datasets, while maintaining stability on the ZJU dataset. Subsequently, HFFM is replaced with our proposed GFRM, with results shown in the third row. Finally, FFRM is replaced with the proposed FEIM, with results presented in the fourth row. The experimental results underscore the individual effectiveness and compatibility of our proposed GFRM, LFFM, and FEIM. When fully integrated, these modules significantly enhance RoadFormer+'s performance in processing three types of RGB-X data compared to the original RoadFormer's feature fusion method. The feature fusion method presented in row four is our proposed MHFF block. To further validate the effectiveness of the channel expansion design in LFFM and the collaborative processing of $\boldsymbol{Z}^h_i$ and $\boldsymbol{Z}^w_i$ in FEIM, additional experiments are conducted. Removing these operations leads to a decline in the overall performance, as demonstrated in rows five and six.

\begin{table}[t!]
\fontsize{7.2}{11}\selectfont
\centering
\caption{
Ablation study on the effectiveness of our proposed MHFF block.}
{
\begin{tabular}{L{3.3cm}|c|c|c}
\toprule[1pt]
Feature Fusion Method & \makecell{SYN-UDTIRI \\ IoU (\%)$\uparrow$}  & \makecell{MFNet \\ mIoU (\%)$\uparrow$}  & \makecell{ZJU \\ mIoU (\%)$\uparrow$}   \\
\hline
HFFM + FFRM & 93.44  & {59.34} & {92.72}  \\
HFFM + LFFM + FFRM  & 93.67 & {60.13} & {92.70} \\
GFRM + LFFM + FFRM  & 93.82 & {60.51} & {92.85} \\
GFRM + LFFM + FEIM  &  \textbf{93.91}  & {\textbf{60.91}} & {\textbf{92.89}} \\
\cline{1-4}
GFRM + LFFM\textsuperscript{\ding{72}} + FEIM  &  {93.45} & {60.69} & {92.64} \\
{GFRM + LFFM + FEIM\textsuperscript{\ding{73}}}  &  {93.76} & {59.42} & {92.55} \\
\bottomrule[1pt]
\end{tabular}
}
\\[5pt] 
\footnotesize{\textsuperscript{\ding{72}} The feature channel number of LFFM is doubled due to direct duplication.
\\[2pt] 
\textsuperscript{\ding{73}} $\boldsymbol{Z}^h_i$ and $\boldsymbol{Z}^w_i$ in the FEIM are processed separately without interaction.}
\label{tab.SFFM_ablation_3}
\end{table}

\section{CONCLUSION}
\label{Sect.conclusion}
This article reviewed designs for heterogeneous feature extraction and fusion strategies and introduced RoadFormer+, a highly efficient, robust, and applicable urban scene parsing network. Breaking down our contributions further, our work contains five key technical advancements: two modules for feature decoupling in the encoding stage, and three new components within the feature fusion module. The effectiveness of each contribution was validated through extensive experiments. RoadFormer+ outperforms other SoTA algorithms across multiple RGB-X scene parsing datasets. Our future work will primarily focus on investigating lightweight algorithms to enhance adaptability to terminal devices.

\bibliographystyle{IEEEtran}
\bibliography{./ref.bib}

\end{document}